\documentclass[letterpaper, 10 pt, conference]{ieeeconf}  

\IEEEoverridecommandlockouts                              

\usepackage{graphics} 
\usepackage{epsfig} 
\usepackage{mathptmx} 
\usepackage{times} 
\usepackage{amsmath} 
\usepackage{amssymb}  
\usepackage[rgb]{xcolor}

\title{\LARGE \bf Soft, Round, High Resolution Tactile Fingertip Sensors for Dexterous Robotic Manipulation}

\author{
   \authorblockN{Branden Romero$^{1}$, Filipe Veiga$^{1}$, and Edward Adelson$^{1}$} %
       \authorblockA{$^{1}$Massachusetts Institute of Technology
   {\tt\small <brromero,fveiga>@mit.edu, adelson@csail.mit.edu}}
\thanks{*This work was supported by the Toyota Research Institute and the Office of Naval Research}%
}
\begin{document}

\maketitle
\thispagestyle{empty}
\pagestyle{empty}

\begin{abstract}
High resolution tactile sensors are often bulky and have shape profiles that make them awkward for use in manipulation. This becomes important when using such sensors as fingertips for dexterous multi-fingered hands, where boxy or planar fingertips limit the available set of smooth manipulation strategies. High resolution optical based sensors such as GelSight have until now been constrained to relatively flat geometries due to constraints on illumination geometry. Here, we show how to construct a rounded fingertip that utilizes a form of light piping for directional illumination. Our sensors can replace the standard rounded fingertips of the Allegro hand. They can capture high resolution maps of the contact surfaces, and can be used to support various dexterous manipulation tasks.
\end{abstract}

\section{INTRODUCTION}

The sense of touch has been shown to greatly contribute to the dexterity of human manipulation, especially in cases where high precision is required~\cite{johanssonvideo}. The complex ensemble of mechanoreceptors in the human hand provides extremely rich tactile sensory signals~\cite{johansson1979tactile}. These sensory signals encode information such as contact force and contact shape, and can be used to detect complex state transitions such as making or breaking contact or the occurrence of slippage between the finger and the grasped object~\cite{johansson2009coding}.

In recent years, vision based tactile sensors have become very prominent due to their high signal resolutions and the softness of their sensing surfaces~\cite{lepora2015superresolution,yuan2017gelsight,yamaguchi2016combining}. The softness of the sensing surface allows for larger contact regions as it deforms to comply with the object surface. The resulting contact areas are then characterized in great detail via the high resolution signals. Together, these properties have enabled the use of these sensors in tackling several tasks such as assessing grasp success~\cite{calandra2018more}, servoing object surfaces~\cite{lepora2017exploratory}, detecting slip and shear force~\cite{yuan2015measurement}, reconstructing 3D surfaces~\cite{wang20183d} and distinguishing between different cloth materials~\cite{yuan2018active}. The high resolution signals provided by the sensors does come at a cost with sensor design being constrained in terms shape~\cite{yuan2017gelsight,yamaguchi2016combining}. Sensors from the TacTip ~\cite{tactip} family have been developed in a wide range of geometries. However, the very high resolution sensors based on GelSight have been constrained to flat or nearly flat designs, due to the difficulties of providing well controlled directional lighting with non-planar geometries. Our goal here is to remove this design constraint and to allow the creation of GelSight fingertips that are rounded.

\begin{figure}[t]
    \centering
    \includegraphics[width=\linewidth]{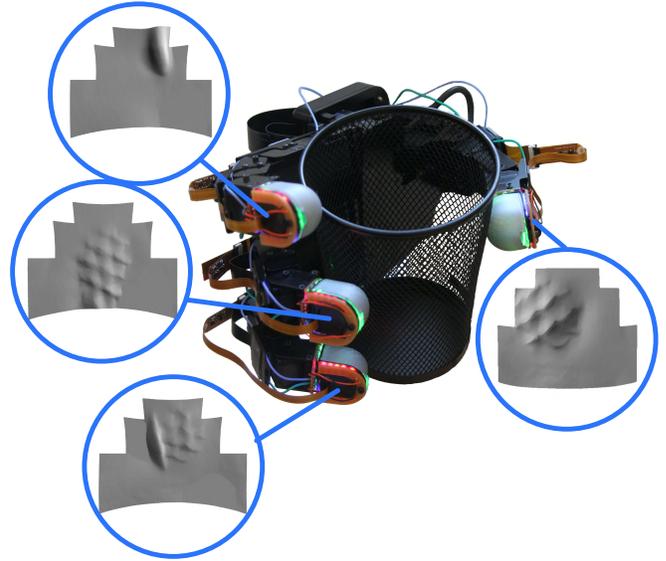}
    \caption{An Allegro Hand equipped with four sensors holding a mesh cup. Each finger provides a high resolution 3D reconstruction of its respective contact areas.}
    \label{fig:my_label}
    \vspace{-0.5cm}
\end{figure}

In this paper we present a GelSight fingertip sensor, where we preserve the softness and high signal resolution signals associated with the classic GelSight sensor~\cite{yuan2017gelsight}, while dramatically changing the sensors shape to better suit the needs for dexterous manipulation task. We begin by providing a discussion on the advantages of a round finger tip as opposed to a flat one (Sec.~\ref{sec:curved_vs_flat}), describing the design choices made to achieve the target shape (Sec.~\ref{sec:design}) and showcasing the manufacturing procedures (Sec.~\ref{sec:manufacturing}). We follow with a description of the sensor calibration, used for recovering 3D surface deformations from the GelSight sensor images (Sec.~\ref{sec:calibration}) and by a description of the method we use to estimate and track the contact areas (Sec.~\ref{sec:contact}). Finally, we present several experiments where we use the sensor signals and take advantage of the sensor's shape to manipulate objects (Sec.~\ref{sec:exp}) and discuss the outcome of our work while also providing some future directions (Sec.~\ref{sec:conclusion}).

\section{DESIGN AND MANUFACTURING}
\subsection{Importance of Shape in Fingertip Sensors}
\label{sec:curved_vs_flat}

\begin{figure}[t]
    \centering
    \includegraphics[width=0.49\linewidth]{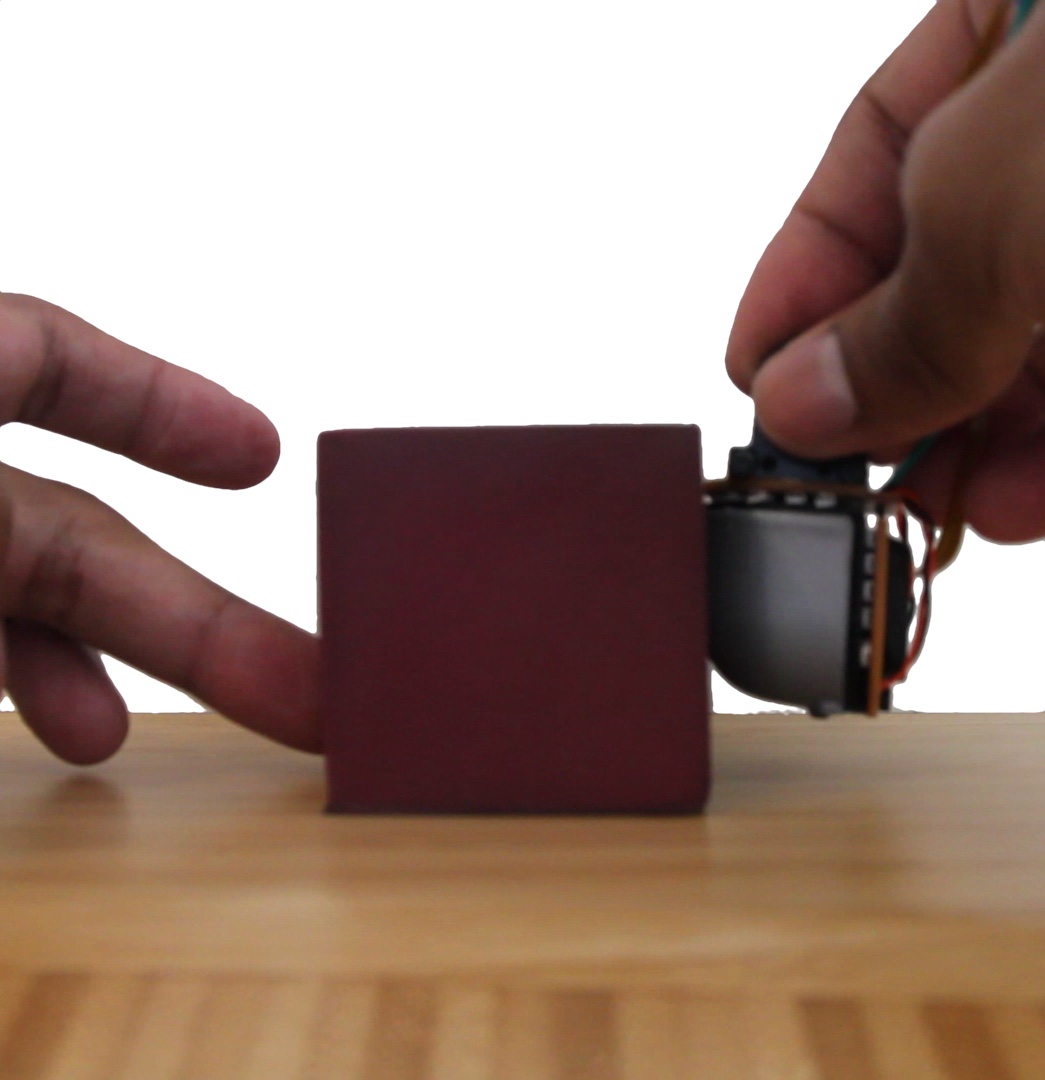}
    \includegraphics[width=0.49\linewidth]{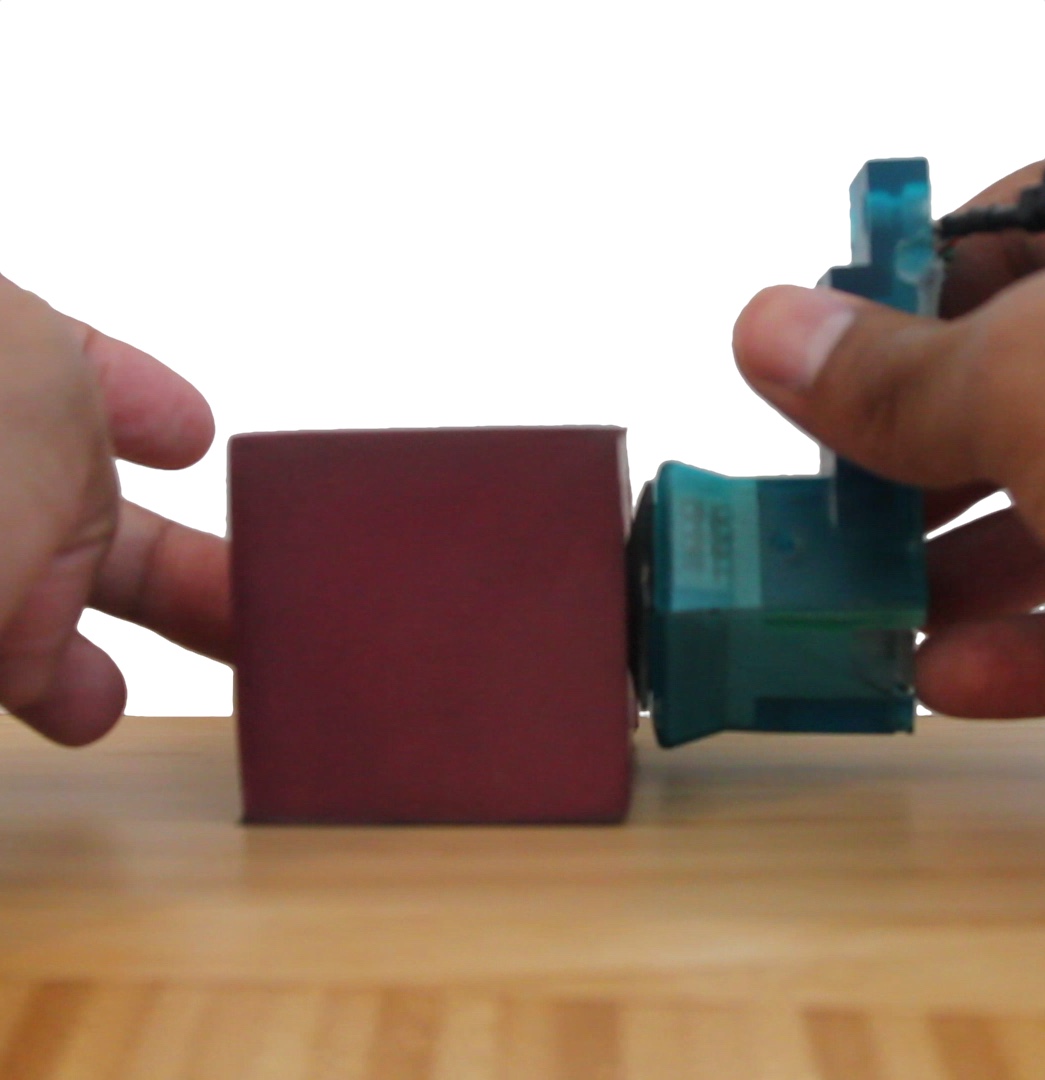}
    \includegraphics[width=0.49\linewidth]{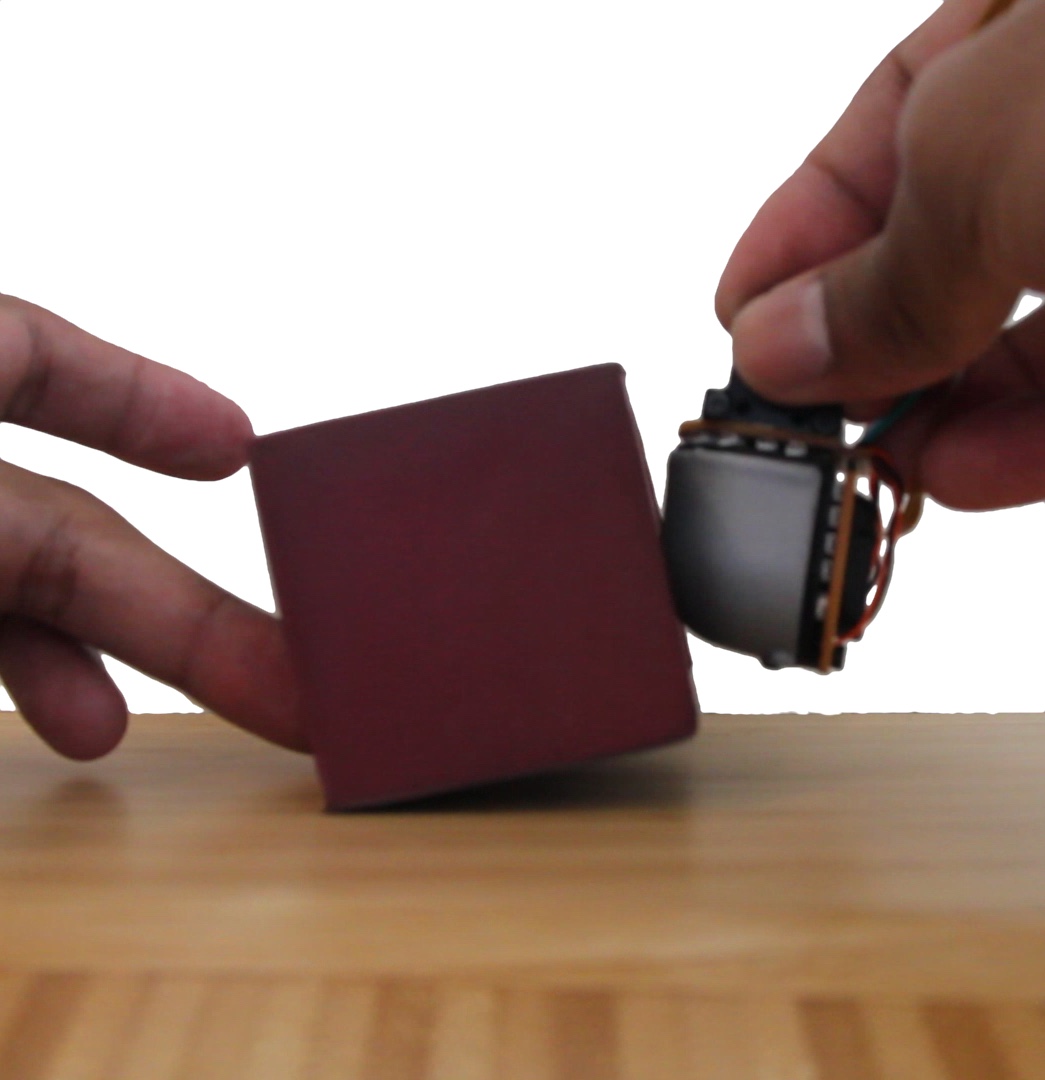}
    \includegraphics[width=0.49\linewidth]{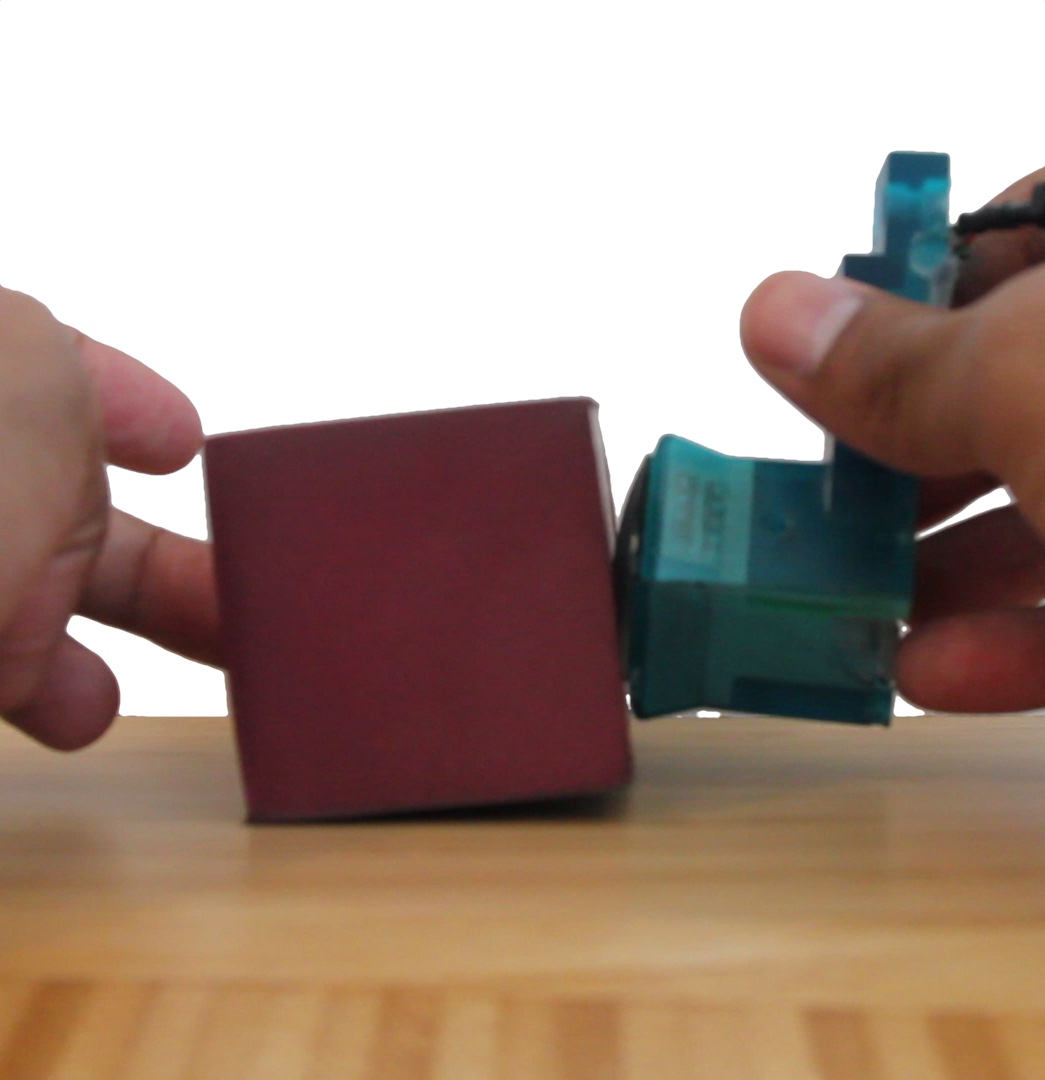}
    \includegraphics[width=0.49\linewidth]{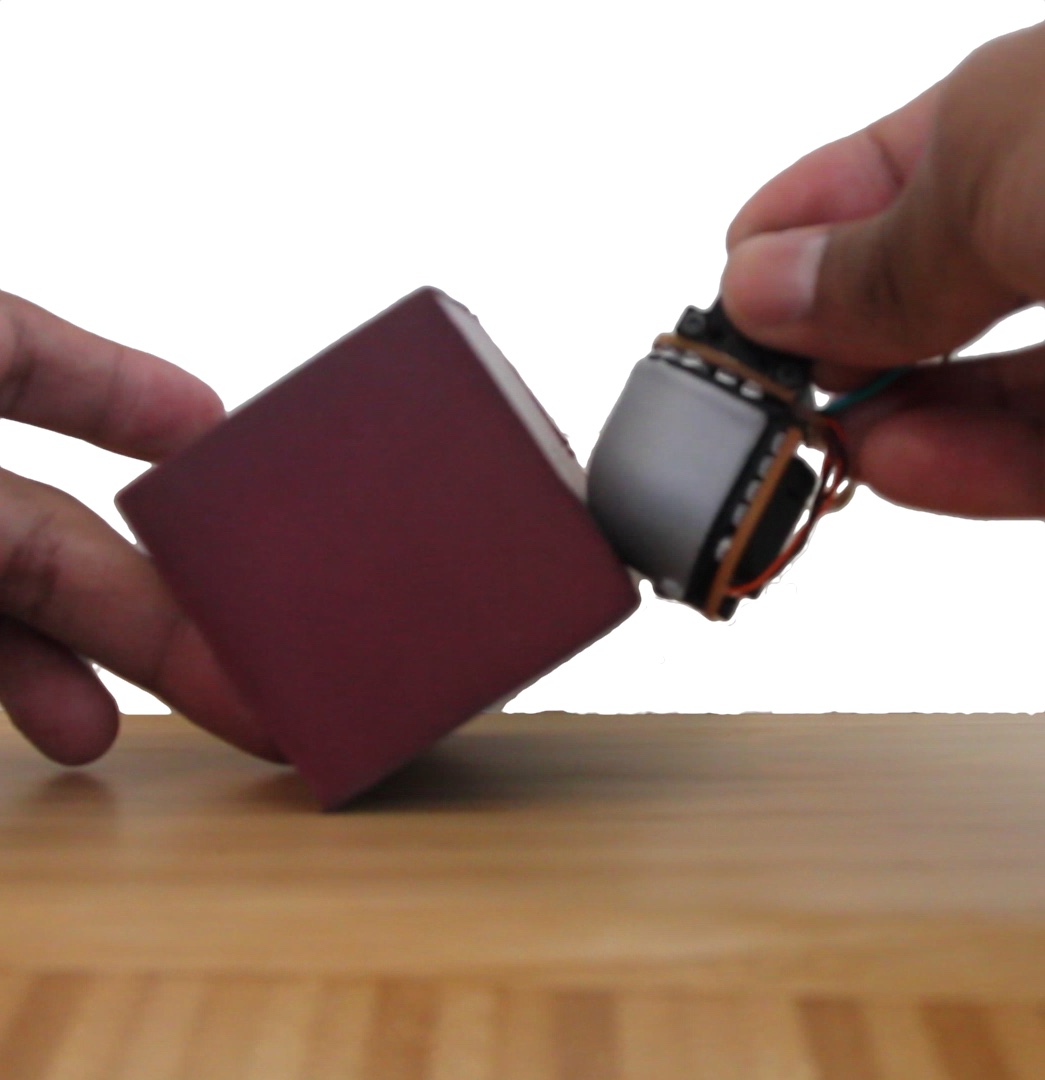}
    \includegraphics[width=0.49\linewidth]{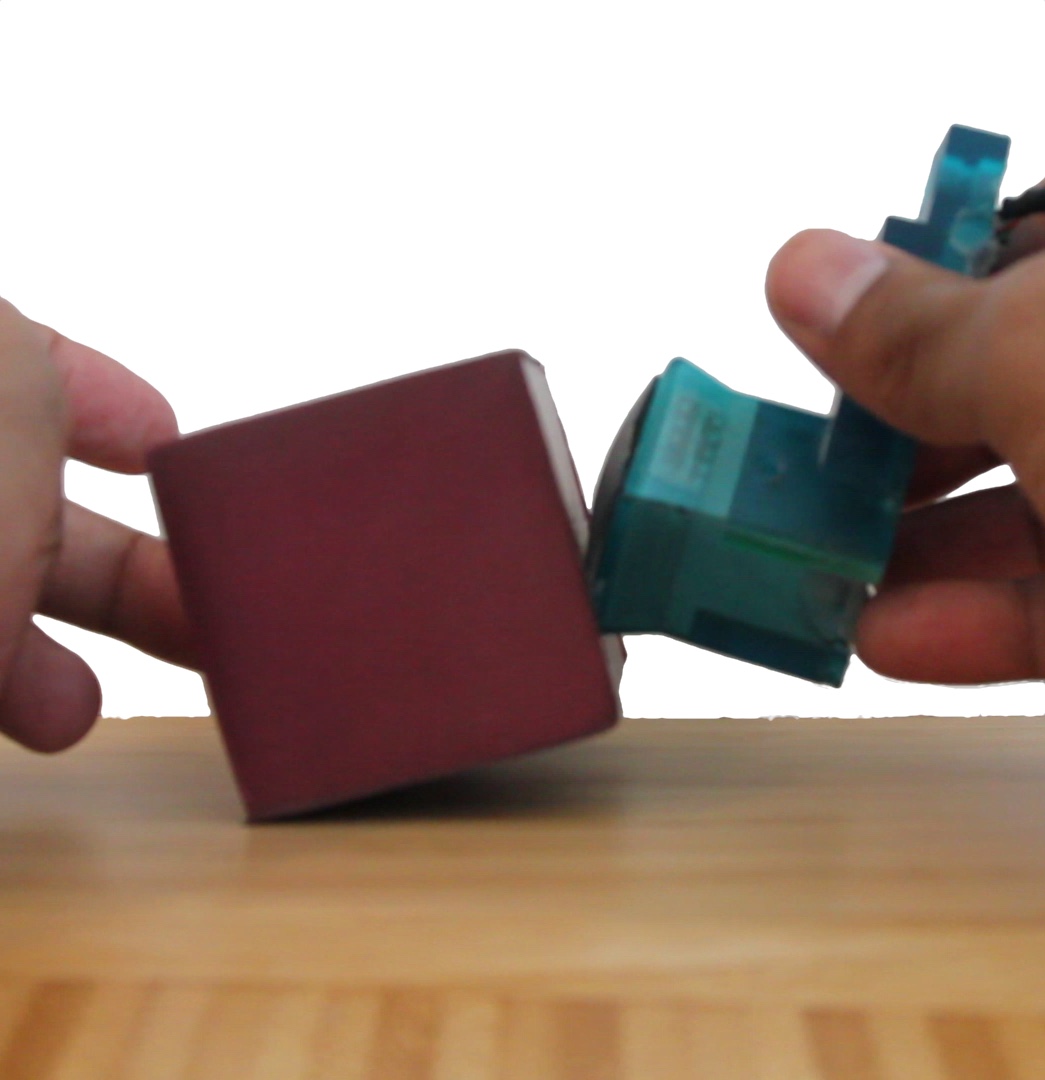}
    \caption{Human demonstration illustrating rolling a cube along the frontal plane of the finger. (a) the rolling motion of a curved finger which maintains a constant contact area size and has a wider range of motion. (b) shows the rolling motion with a flat finger which has a shrinking contact area and limited range of motion. }
    \label{fig:curved_vs_flat}
    \vspace{-0.5cm}

\end{figure}

\begin{figure}[t]
    \centering
    \includegraphics[width=0.49\linewidth]{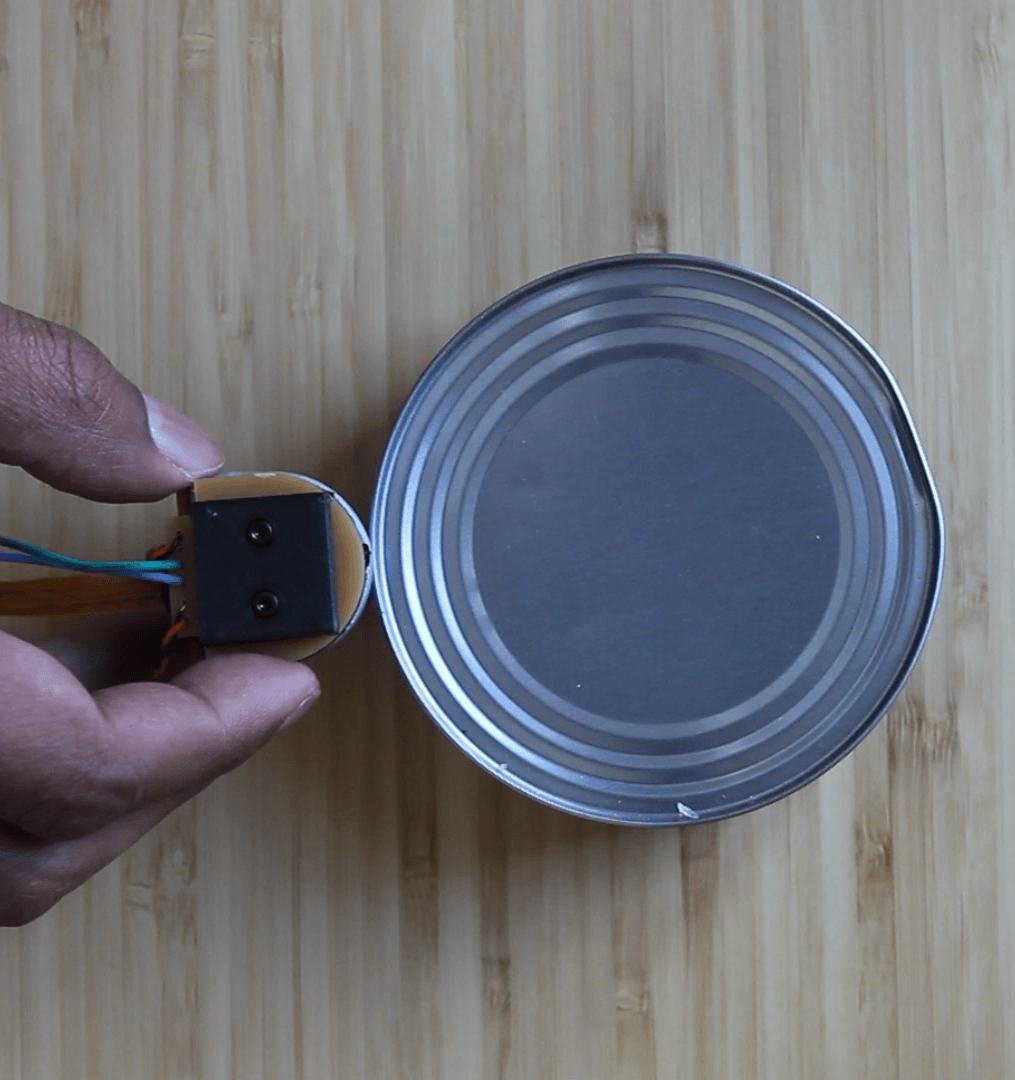}
    \includegraphics[width=0.49\linewidth]{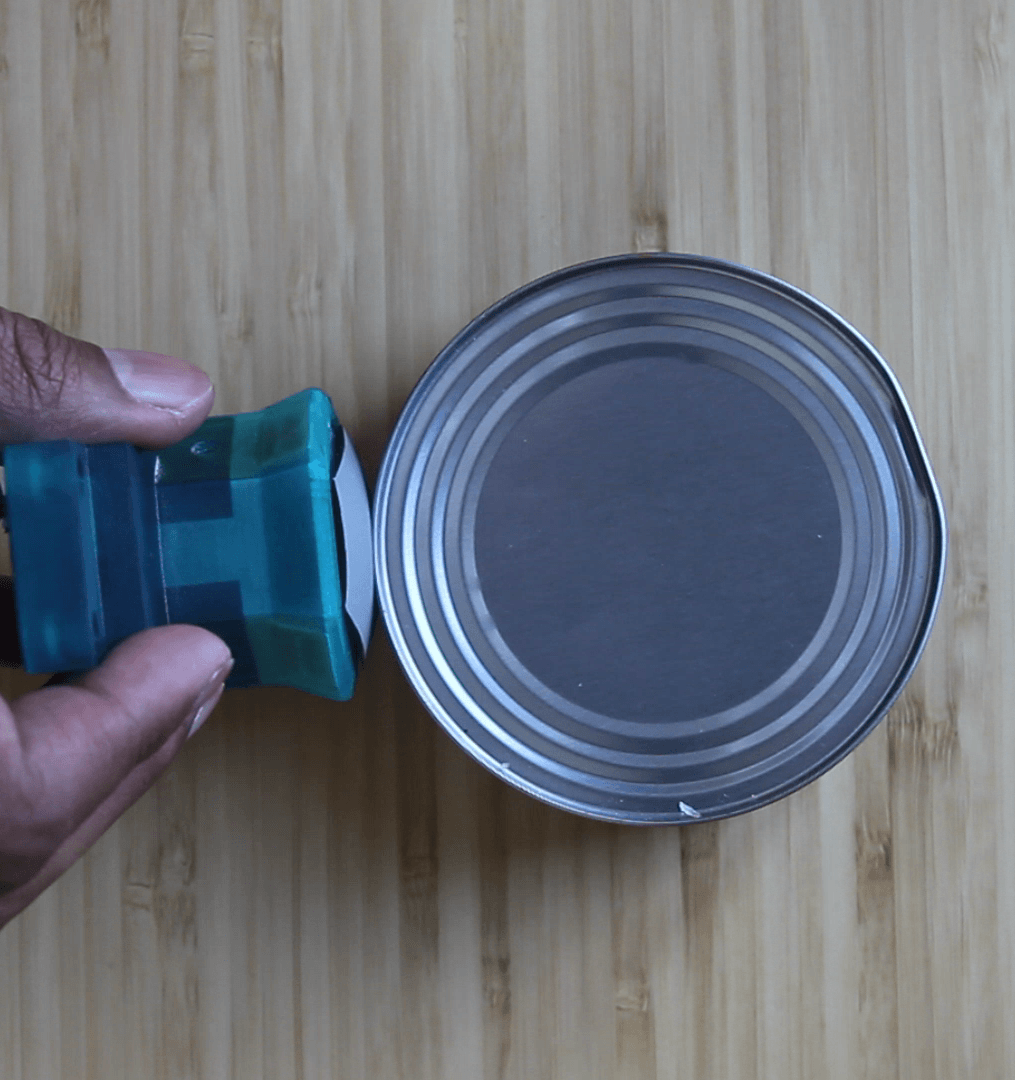}\vspace{0.2cm}
    \includegraphics[width=0.49\linewidth]{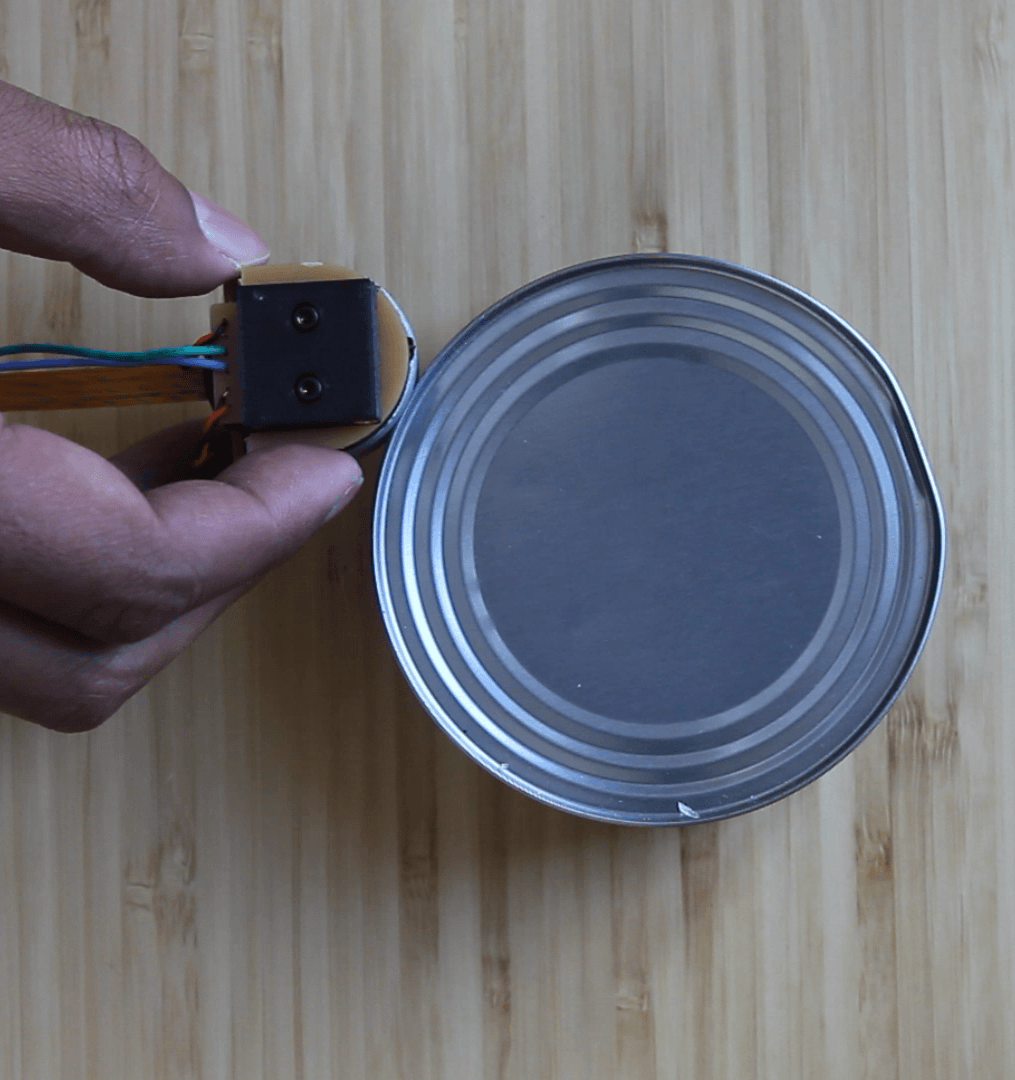}
    \includegraphics[width=0.49\linewidth]{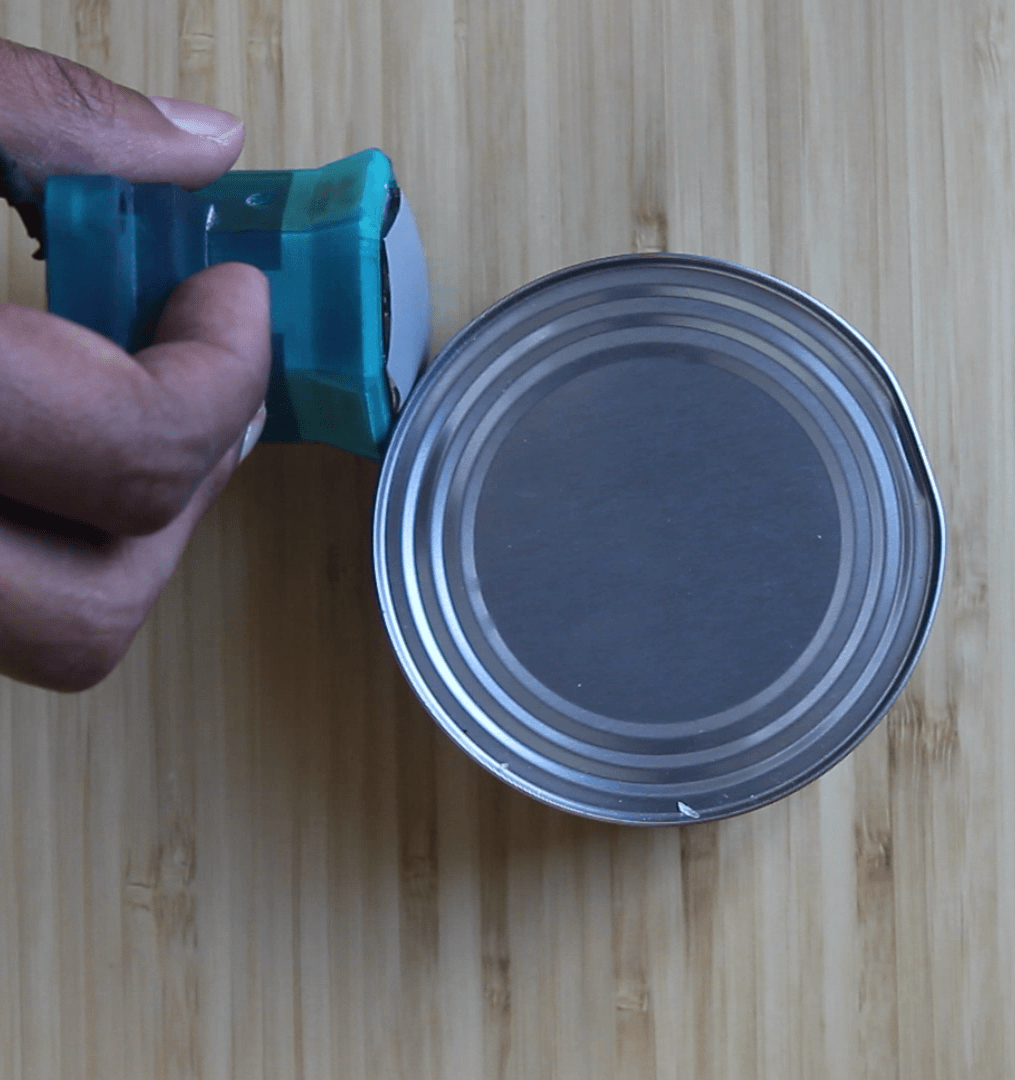}
    \caption{Human demonstration illustrating contacts made on different parts of a cylinder using the finger's transverse plane. (a) the contact areas made with a curved finger,(b) the contact areas made with a flat finger. The round finger keeps a consistent contact area at each contact point unlike the flat finger. }
    \label{fig:curved_vs_flat2}
    \vspace{-0.5cm}

\end{figure}

Consider the task of flipping an object that is resting on a table, as depicted in Fig.~\ref{fig:curved_vs_flat}. Here the index finger of a hand rolls an object that is lying on a supporting surface towards the thumb. Of the two cases depicted, we can see that when using flat sensors, the contact patch location and size greatly vary throughout the manipulation trajectory, with the size of the contact patch being reduced to almost a point contact when it reaches the edge of the sensor. Having such a small contact patch not only reduces the stability of the object, but also reduces the robots perception of the objects state. On the other hand, when a curved sensor is used, while the location of contact still changes in a similar manner, the contact patch size remains relatively consistent throughout the object trajectory. This decrease in variation of the contact patch size makes it much easier to track the object state.

Another case where fingertip shape clearly impacts the performance of dexterous hand systems is when performing grasp quality assessment. For assessing the quality of a grasp it is critical that the contact areas acquired after grasping the object are perceivable by the fingertip sensors. When using flat sensors, in order to maximize the contact patch information, the fingers have to be reoriented such that the sensing surface is orthogonal to the contact location. As previously stated, this can be problematic when considering the limited kinematic structure of each finger. Since curved fingertips are able to perceive contact patches in a wider ranged of orientations, explicitly reorienting the fingertip becomes unnecessary. An example of a grasp configuration where the differences between the two sensors are visible is also depicted in Fig.~\ref{fig:curved_vs_flat2}.

\subsection{Design}
\label{sec:design}

\begin{figure}[t]
    \centering
    \def\svgwidth{\linewidth}
    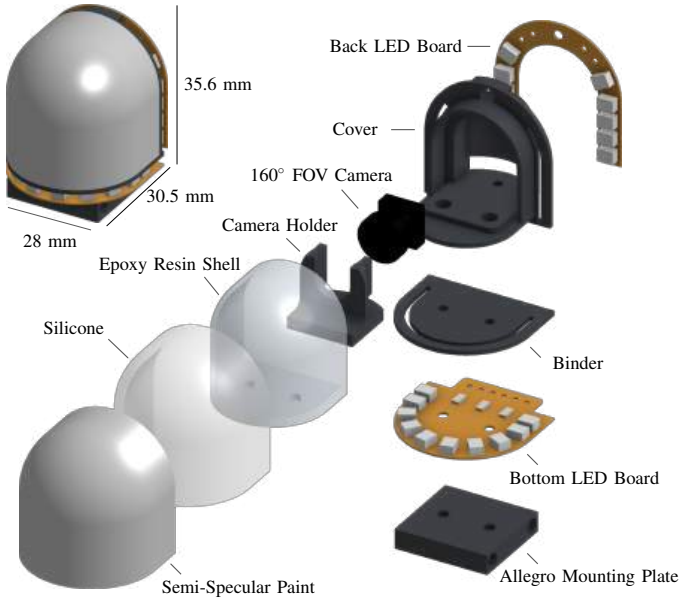
    \caption{CAD model of the sensor illustrating the assembled finger, along with the exploded view of the sensor showing the internal components.}
    \label{fig:assembly}
    \vspace{-0.5cm}

\end{figure}

The goal of our design (Fig.~\ref{fig:assembly}) is to enable robots with dexterous manipulators to have rich information about the contact, in particular dense information about the 3D geometry of the contact areas. We propose a novel illumination system, that despite more complex geometry, allows us to perform 3D height map reconstruction despite some limitations.

To achieve this design we make sacrifices in terms of illumination quality as seen in the previous sensor ~\cite{yuan2017gelsight} hoping that the reconstructed height map is still suitable for robotic manipulation. In particular we use an opaque sensing surface that does not have a one-to-one mapping between surface normals and RGB values, and secondly while the previous sensor provides information about the x-gradient and the y-gradient from at least two sources of information via direction lighting, we only provide information about the x-gradient with two sources of information and the y-gradient with one source of information.

\begin{figure}[t]
    \centering
    \includegraphics[width=\linewidth]{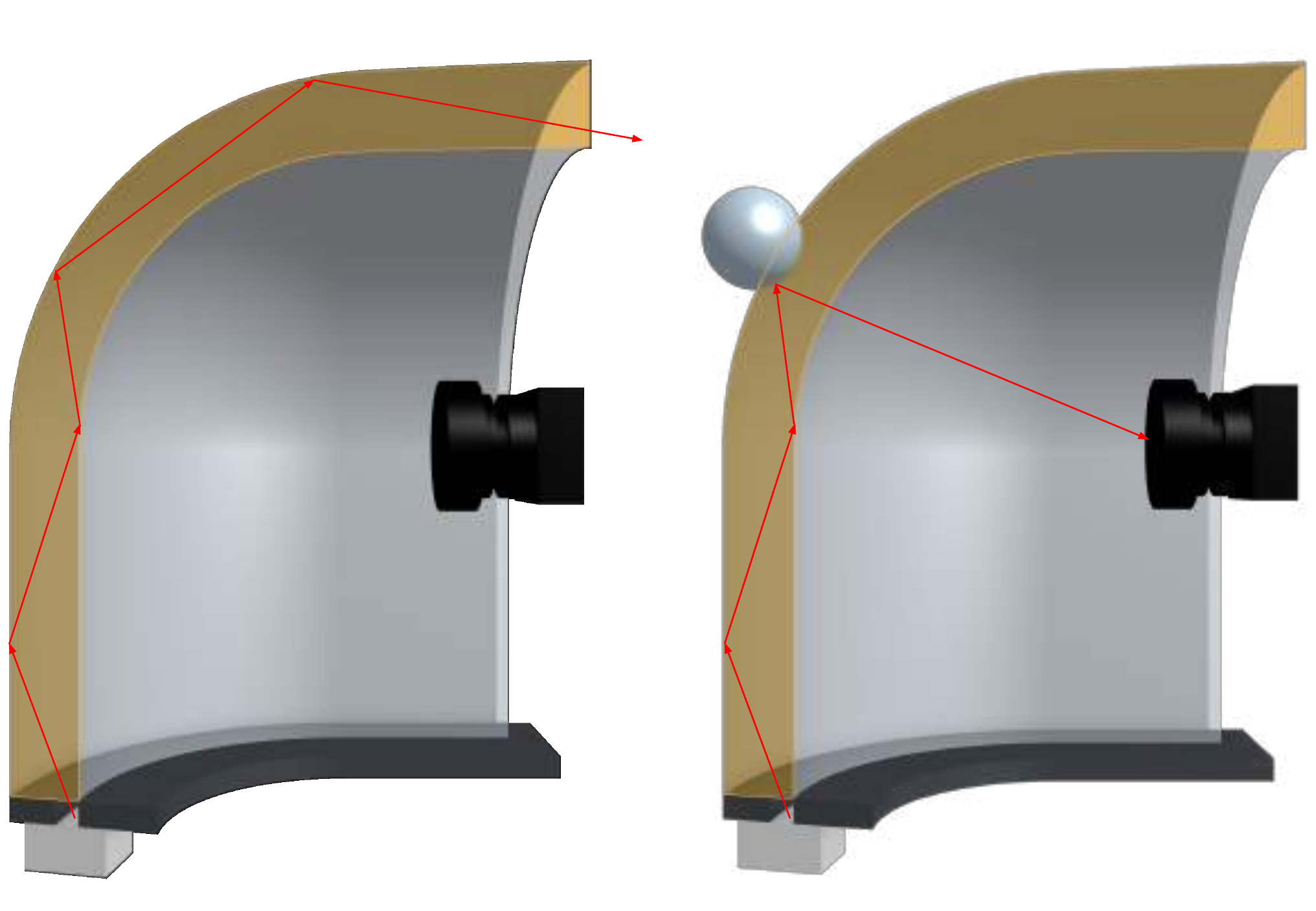}
    \caption{Illustration showing the light piping illumination system along a single axis. (a) the path in which the light travels without contact. (b) the path of the light when contact is made.}
    \label{fig:light_piping}
    \vspace{-0.5cm}

\end{figure}

These choices were made in order to provide as uniform of an illumination pattern as possible along the entire surface of the finger. To do this we relied on a technique called light piping. Light piping is inspired by fiber optics in which a light is constrained within a medium via total internal reflection (TIR). We achieve something similar by using a semi-specular sensing surface and a thin plastic shell. We design the sensor in this manner, because semi-specular surfaces only slightly diffuses light while the lambertian surface used in the previous sensor completely diffuses the light. This means that rather than the light dissipating as it travels along the sensor like it would with a lambertian surface, it will keep its directionality and thus more uniformly illuminate the surfaces beyond the curve of the surface Fig.~\ref{fig:light_piping}. The thin plastic shell keeps the light from escaping into the interior of the sensor via TIR unless contact is made.

\subsection{Manufacturing}
\label{sec:manufacturing}

As a consequence of a more complicated geometry we have a more complicated manufacturing process compared to previous versions of the GelSight sensors. Where the previous sensor only required 3D printing, laser cutting, casting silicone into an open face mold, and pour over methods for coating, we rely on manufacturing a series of two-piece molds, casting into said molds, and then applying the opaque coating using an airbrush. However, in this process we have created methods that also significantly increase the durability and reliability of the sensors compared to the previous version.

\subsubsection{Mold Making}
First, the desired geometry of the plastic shell, as well as the geometry of the silicone and the shell combined, are 3D printed with the Form Labs Form2 SLA printer using the clear resin. The cast pieces have to be optically clear, so we prime the 3D printed pieces with Krylon Crystal Clear and then dip the pieces in a clear UV cure resin and let it drip until a thin layer remains. We then cure the resin with a UV lamp. We do this multiple times until the print is smooth and optically clear. The reference plastic shell piece is then used to create a two-piece silicone mold. We chose Smooth-On MoldMax XLS II as our mold material because we will cast epoxy resin into these molds later, and we found this material robust to multiple castings compared to other mold materials we used.

The second mold will be used to cast silicone onto the plastic shell, so we create a two piece mold where the base of the mold is a rigid 3D printed piece that will be rigidly attached to the shell, and the other piece is a soft silicone mold made from Smooth-On MoldStar 20T.

\begin{figure*}[t]
    \centering
    \def\svgwidth{\linewidth}
    \input{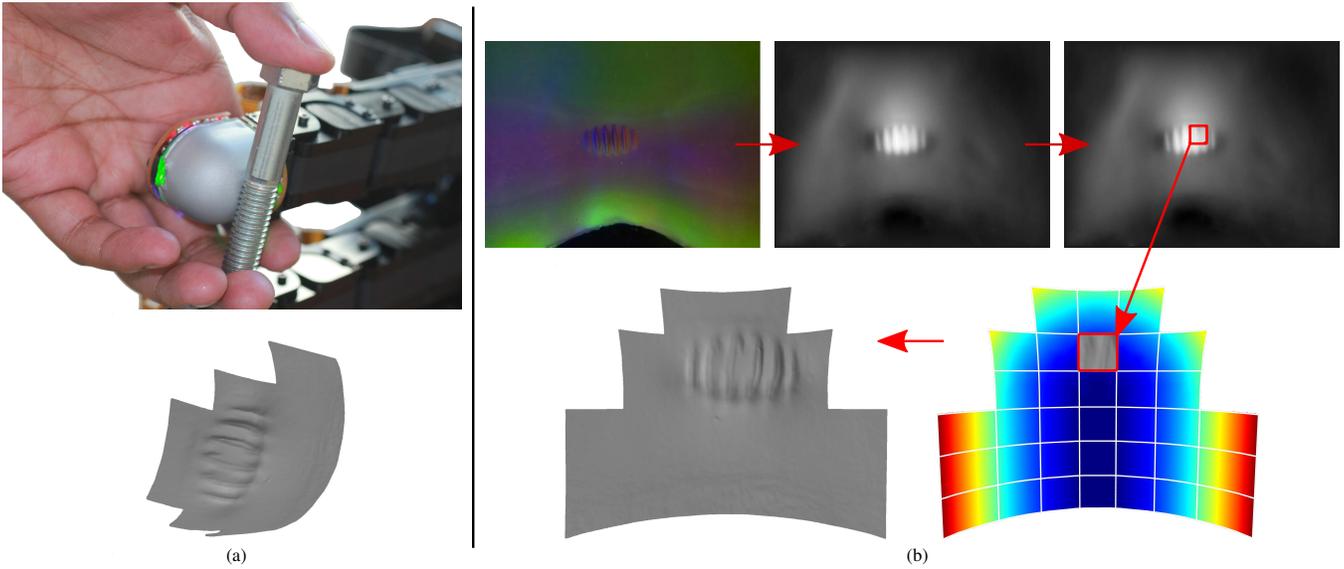}
    \caption{(a) the \textbf{bottom} shows the 3D reconstruction from the contact made textbf{top} from pressing the screw on the sensing surface. (b) Shows the 3D reconstruction pipeline. \textbf{Top from left to right} shows the raw image, the height map generated from the fast Possion solver, and the image patch that is used to generate \textbf{bottom right} the point cloud patch. \textbf{Bottom left} image show the complete 3D reconstruction. }
    \label{fig:data_pipeline}
    \vspace{-0.5cm}
\end{figure*}
\subsubsection{Casting}
We begin the process by casting the plastic shell. Here we chose to cast the shell with a clear epoxy resin, in particular Smooth-On Epoxacast 690. The shell is only 1mm thick so we chose this material because of its low-viscosity, clarity, rigidity, and overall ease of use. After casting the shell we let the piece sit for 24 hours to completely cure.
The next part tries to address some of the limitations of the previous sensor in terms of durability. In particular, in the previous sensor the paint was easy to remove and the gel easily delaminated from acrylic window. These issues stem from the fact that, one it is difficult to get silicone to stick to anything, and secondly that things were mechanically attached rather than chemically. To start we begin by priming the surface of our plastic shell with Dow DOWSIL P5200. This promotes silicone's adhesion to a variety of surfaces, but the caveat is the silicone must cure on that surface to form a chemical bond, rather than mechanically attaching a cured piece of silicone to a surface like in the previous Gelsights. The next part is creating the opaque coating. After experimenting with a variety of coatings we decided on creating a custom coating that is quite durable. The coating is made out of a silicone paint base, Smooth-on Psycho Paint, and a non-leafing silver dollar aluminum flake pigment. We spray this coating, so we dilute it with a silicone solvent, Smooth-On NOVOCS. The exact ratio used is 1:10:30 pigment, silicone paint base, and silicone solvent ratio by mass. We spray the interior of the silicone part of our mold with Mann Release Technologies Ease Release 200, and then spray our opaque coating in with an airbrush. We quickly screw our plastic shell onto our mold base, assemble the mold, and pour our optically clear silicone gel (Silicone Inc. XPS-565 1:15 A:B ratio by mass) into the mold. We want to cast the silicone before the coating cures so that they are chemically bonded. The mold is left out for 6 hours at room temperature and then is placed in the oven at 95 degrees Celsius.

\begin{figure}[b!]
    \centering
    \includegraphics[width=\linewidth]{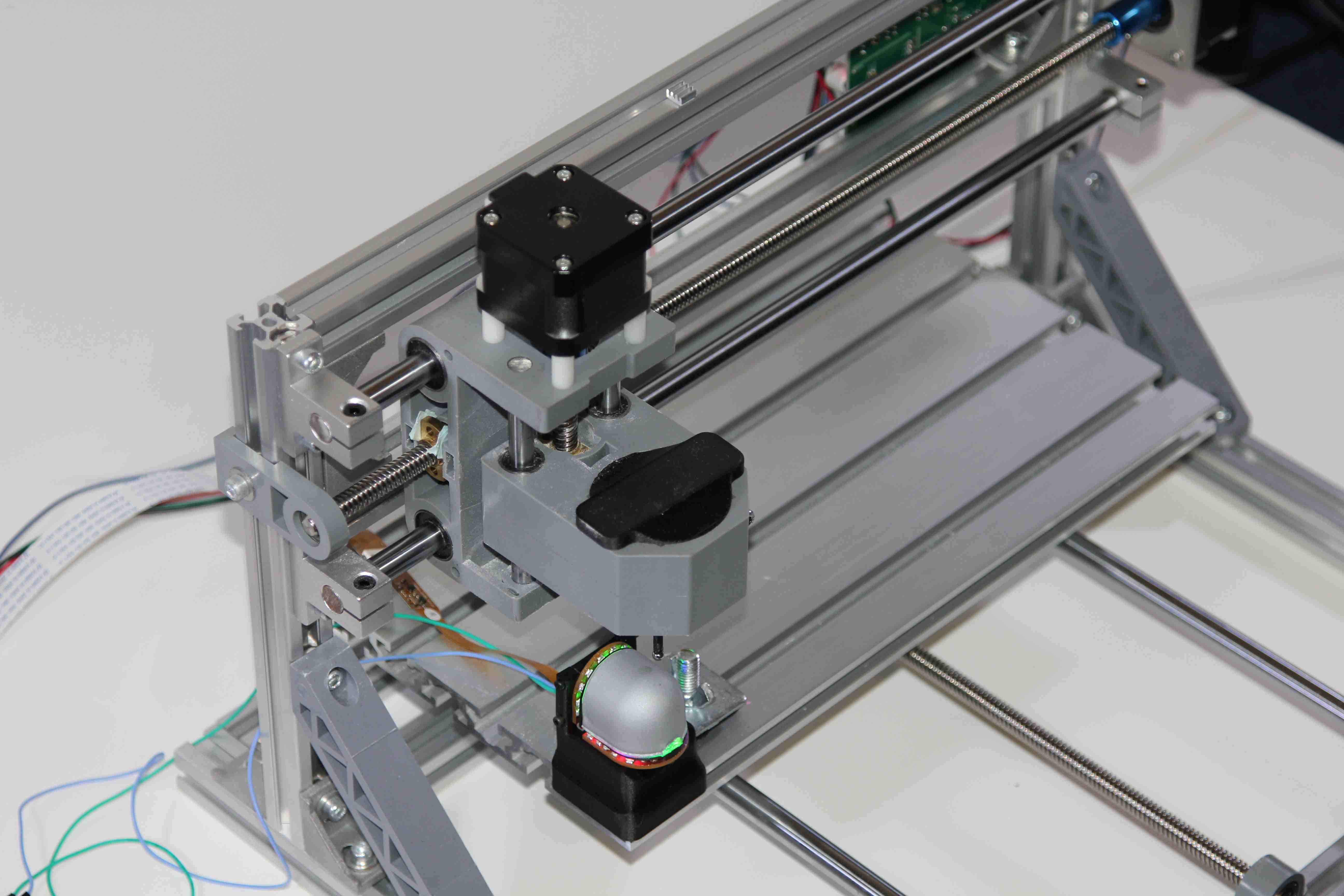}
    \caption{The sensor attached to the CNC rig used for calibration. The sensor pokes the surface of the sensor at a variety of locations to construct the look-up table for 3D reconstruction and map the 3D surface to a 2D image.  }
    \label{fig:cnc}
\end{figure}

\subsubsection{Assembly}
The camera holder, cover, blinder, and mounting plate are 3D printed on the Markforged Onyx One printer with the Onyx filament which is very suitable for creating strong fixtures. The camera, Frank-S15-V1.0 Raspberry Pi camera sensor, is then press fitted into the camera holder. The camera has a high FOV of 160 degrees which observes a significant area of the sensing surface, while being significantly more compact than previous cameras used in GelSight. The camera holder is then press fitted into the cover and then screwed in with a M2 screw.The back LED board is screwed into the cover using an M2 screw. The cover is then press fit into the plastic shell.  Two M2 screws are inserted into the bottom of the Allegro mounting plate that then pass through the through holes of the bottom LED board, blinder, and plastic shell and then screwed into the cover. The two LED boards are soldered together via four wires. Two power cables are routed to a Raspberry Pi 4 to power the LEDs with 3.3V, and then the camera is connected to a Raspberry Pi via CSI flat connector cable. 

\begin{figure*}[t]
    \centering
    \includegraphics[width=\linewidth]{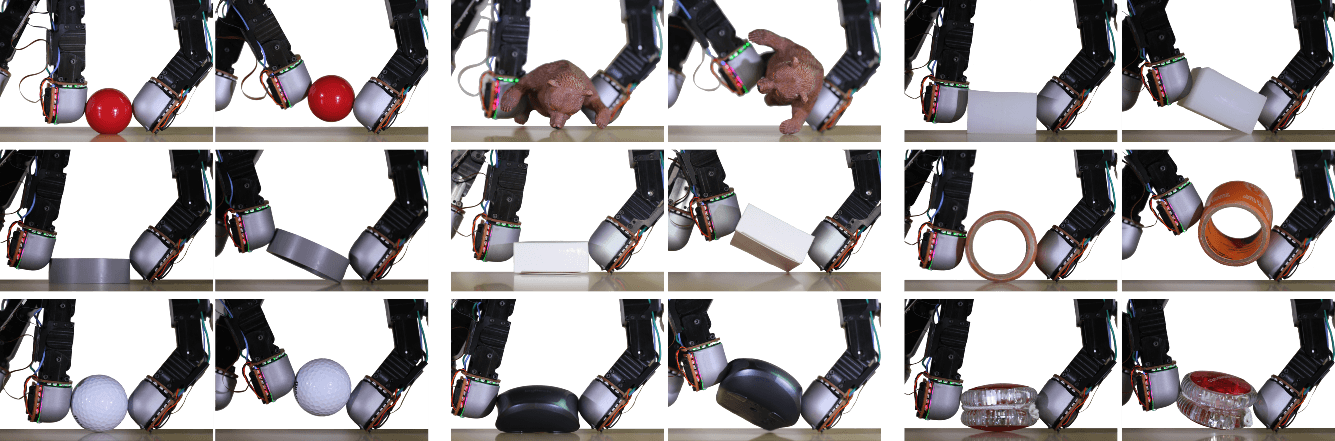}
    \caption{Sample begin and end state of each object during the rolling phase of the experiment.}
    \label{fig:exp}
    \vspace{-0.5cm}
\end{figure*}

\subsection{Software Interface}
The image from the sensor is streamed via HTTP. The image is streamed 640x480 with a frame rate of 90FPS. The latency is low with a measured latency of about 40ms delay. So far no computing happens on the Raspberry Pi. In terms of interfacing all four sensors with a host computer, each camera is connected to a Raspberry Pi 4 using the standard CSI flat connector cables, then the four Raspberry Pis are then connected to a gigabit Ethernet switch in which the host computer is also connected to.

\subsection{Sensor calibration}
\label{sec:calibration}
The calibration process sets out to solve two things. The first is to map RGB values to gradients; the second is to find the 2D-3D correspondence in the form of pixel to point in point cloud correspondence. In terms of mapping RGB values to gradient we can no longer use a single look-up table like the previous GelSight sensors. Despite trying to achieve uniform lighting throughout the whole sensing surface there are obvious non-uniformities, so we propose constructing a look-up table for a set of regions. Once the look-up tables are constructed and we perform 3D reconstruction, the methods know nothing about the curvature of the finger so we propose a piece-wise forward projection of our height map onto the geometry of the surface.

\subsubsection{2D-3D Correspondence}
To begin we start by getting the 3D geometry of our sensing surface. We break up the surface into a set of quads and get the vertices of those quads. We now have to discover where these vertices lie in image space. In order to do this we constructed a CNC rig (Fig.\ref{fig:cnc}) in which we attach the finger tip rigidly in a known location. We then have a probe with a 4mm diameter sphere attached to it. We tell the CNC to poke the sensor at the calculated vertices. On each poke we take a picture of the sensing surface from the sensor’s camera and perform Hough Circle Transform ~\cite{ballard1981generalizing} to find the centroid of the sphere in image space. That point is then added to a table with its corresponding location in the surface’s 3D space. Once all vertices have been probed we construct the reference point cloud. For each quad, we take the corresponding image patch and calculate the perspective transform, since the image is taken from a perspective view. We then warp the image patch and get its resolution. Back in the surface’s space we create a linearly spaced grid in the quad with the same resolution as the image patch and project it onto the surface geometry. Now when we receive a height map image during reconstruction we just take each image patch, warp it, and then change the depth of the corresponding points based-off the depth at each pixel.

\subsubsection{RGB to Gradient Correspondence}
We construct a look-up table mapping each RGB value to a gradient for each image patch determined in the previous section. To construct the gradient, we begin by poking the finger tip in each quad several times at varying locations. For each poke we calculate the centroid and radius of the poke in image space, again using Hough Circle Transform. Since the geometry of the probe is known we can map each pixel intensity to a gradient. For each quad we take the average pixel intensity of all the pokes in that region and map that to the gradient. For RGB values not in the look-up table we assign it a gradient by linearly interpolating the gradients mapped to the nearest RGB values.

\begin{figure*}[t]
    \centering
    \includegraphics[width=\linewidth]{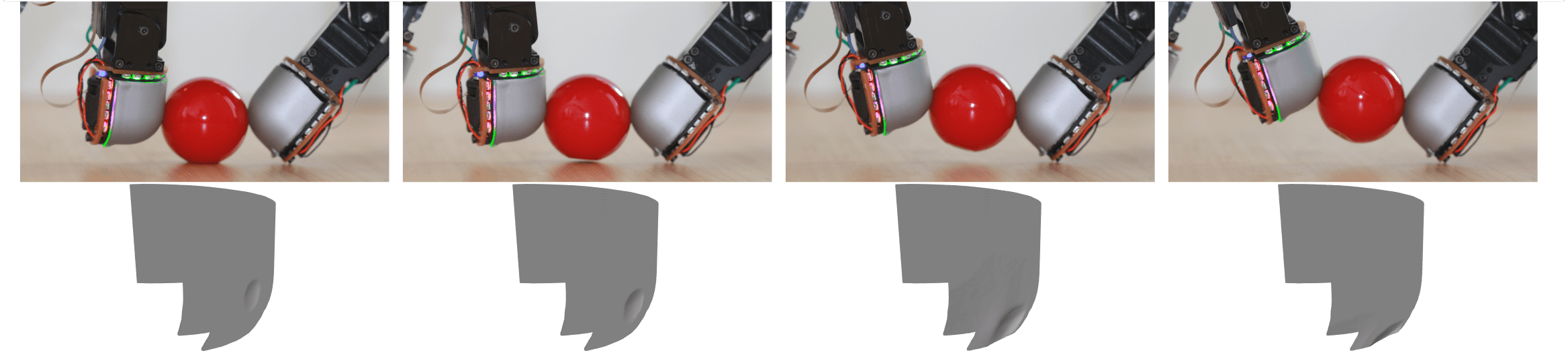}
    \caption{\textbf{Top from left to right} Sequence of the experiment being performed on the plastic sphere during the rolling stage. \textbf{Bottom from left to right} The corresponding point clouds showing the evolution of the point cloud as the object rolls. }
    \label{fig:ball_recon}
    \vspace{-0.5cm}
\end{figure*}

\subsection{Controlled Rolling}
\label{sec:control_roll}
To validate the sensors and the sensor geometry we will perform controlled rolling of a set of unknown objects. In order to do controlled rolling of an unknown object, we propose a tracking method along with a reactive controller to deal with uncertainty in the geometry and dynamics of the object. We assume that no slip will occur throughout execution of the trajectory and chose an action according to the changes in geometry of the sensing surface.

\subsubsection{3D Reconstruction}
In order to achieve reconstruction we need to have fast enough feedback about the geometry of the sensing surface.  Using the tables constructed from the calibration procedure we can now perform real time 3D reconstruction as shown in Fig.~\ref{fig:data_pipeline}. At each time step we create a difference image between the current sensor reading and an image of the sensor without contact to filter out everything in the image except the contact area. We convert the RGB values in the difference image to gradients using our look-up tables. This is passed to our fast Possion solver and we get a height map. Once we receive the height map we extract each image patch corresponding to each quad, warp the image to get rid of the perspective view and add the calculated height to the corresponding point in the point cloud. Our 3D reconstruction runs at 40hz when the image is down-sampled to 320x240.

\subsubsection{Determining Contact Area}
As mentioned in Sec.~\ref{sec:design}, one limitation of the sensor is that we only provide information about the gradient in the x-direction with two sources of information and the gradient in the y-direction with one source of information. This results in inaccuracies about the heightmap along the y-axis. So, basic thresholding on the depth of the heightmap does not result in an accurate contact patch. To address this issue, the contact area is determined by selecting the points with a subset of points with the largest displacement.

\subsubsection{Tracking Contact Area}
To track the contact area we use Iterative Closest Point (ICP) ~\cite{icp}. On each time-step we calculate the current contact area and calculate its convex hull. We get the points from the previous contact area that lie within the convex hull of the current contact area and perform ICP to get the change in the contact area. We only perform ICP with the points within the convex hull since on each time step while rolling we lose contact with areas of the previous contact area.
\label{sec:contact}
\subsubsection{Controller}
We use a hybrid velocity/force controller ~\cite{hybfvc}. This allows the finger to perform a compliant motion where the finger moves up in task space while maintaining a consistent force normal to the the contact patch, resulting in a rolling motion. The maximum displacement of the contact surface is used as a proxy for force in the controller.

\section{RESULTS}
\label{sec:exp}
\subsection{Experimental Setup}
\begin{figure}[t]
    \centering
    \includegraphics[width=0.32\linewidth]{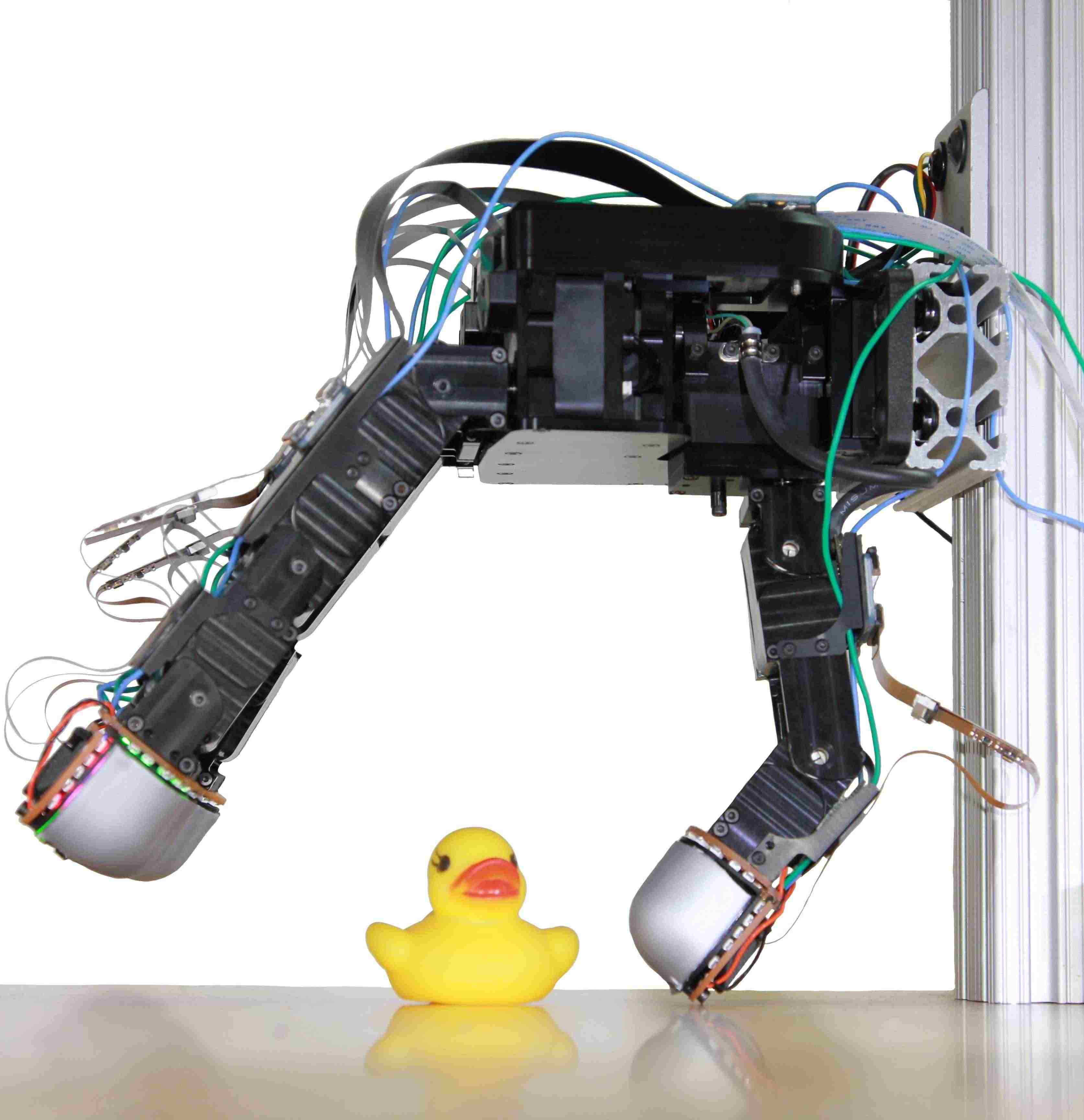}
    \includegraphics[width=0.32\linewidth]{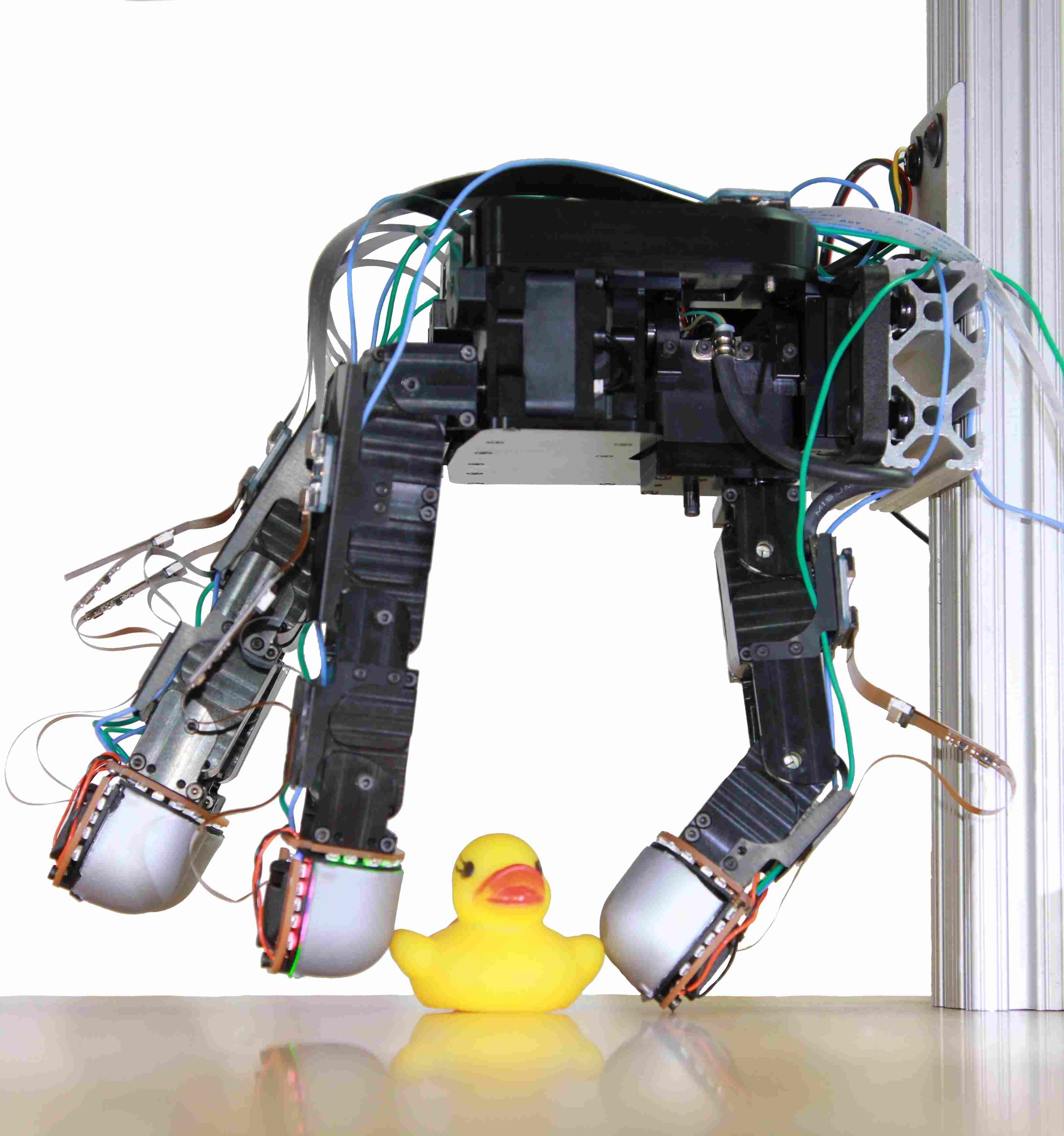}
    \includegraphics[width=0.32\linewidth]{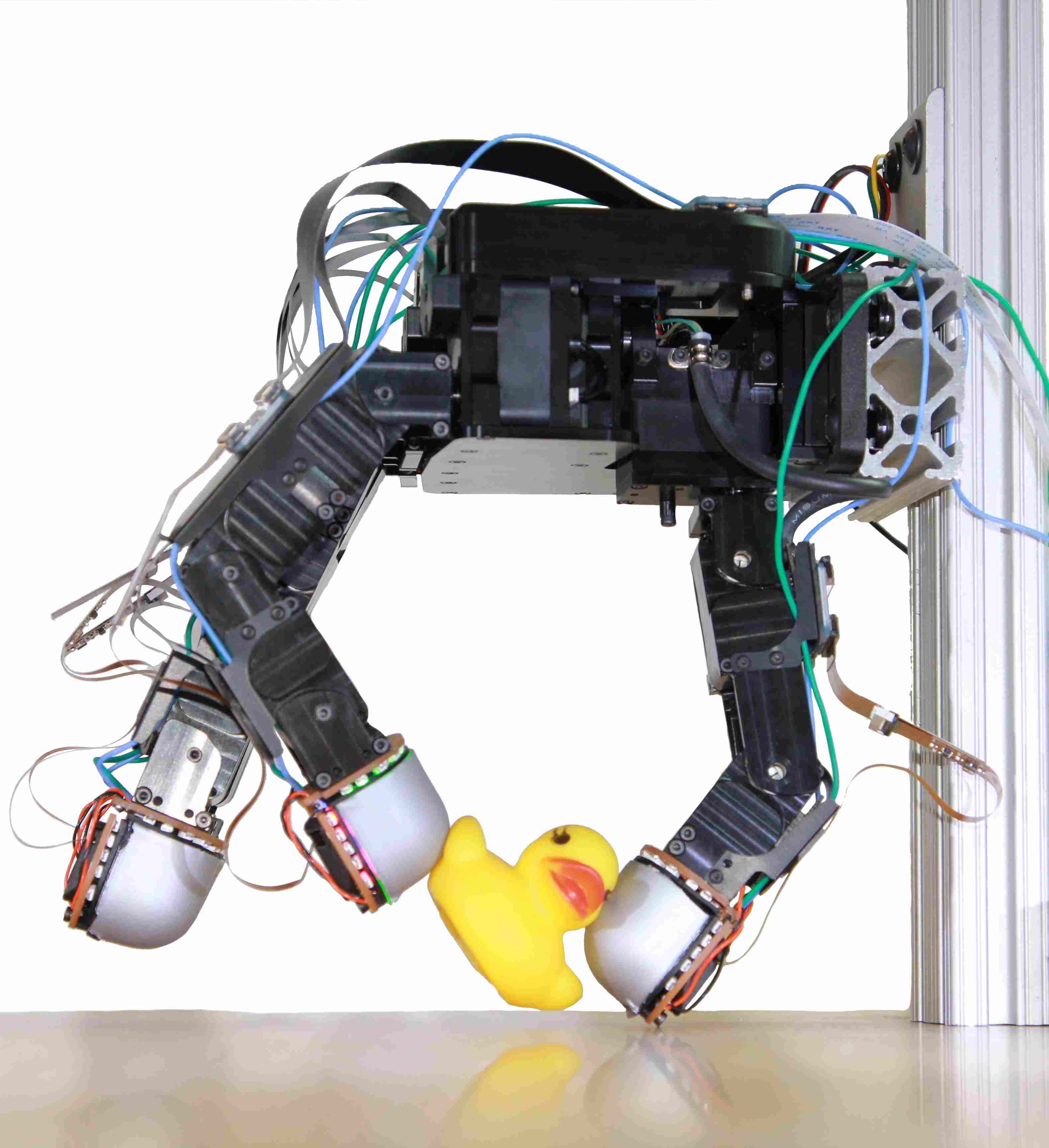}
    \caption{Illustration showing the experimental set-up. \textbf{From left to right} The placement of the object, when the object makes contact with both fingers, the end state of the object when the object reaches the desired position.}
    \label{fig:exp_setup}
    \vspace{-0.5cm}

\end{figure}

To validate our sensor’s capabilities we perform controlled rolling on a set of unknown objects as shown in Fig.~\ref{fig:exp}, Fig.~\ref{fig:ball_recon}, and Fig.~\ref{fig:exp_setup}. The experiment is broken up into three stages. In the first stage we manually place an unknown object in between the index finger and the thumb of the Allegro Hand. While the thumb of the Allegro Hand stays stationary the index finger moves towards the thumb until contact is made. After contact is made, the index finger will continue to move towards the thumb until the desired maximum displacement of the sensing surface is achieved. We then use the controller described Sec.~\ref{sec:control_roll} to roll the object until contact is made in a desired region of the finger as shown in Fig.~\ref{fig:exp_setup}. A trial is considered a success if the finger is able to roll the object until contact is made within the target contact region. If the object falls out of the grasp at any stage in the trial, overshoots the desired contact region, or does not reach the contact area within 3 seconds after the rolling stage of the trial begins the trial is considered a failure. We perform 10 trials for each object.

\subsubsection{Experimental Results}
As illustrated in table (add table here) our method and sensor were able to successfully perform 99 out of 100 controlled rolls into the desired contact region despite being presented a diverse set of objects with varying smoothness, hardness, and geometries. We show a sample of each object being rolled in Fig.~\ref{fig:exp}, Fig.~\ref{fig:ball_recon}, and Fig.~\ref{fig:exp_setup}. Looking at some of the objects, in particular the roll of tape and yo-yo, the object was successfully rolled to the desired position but exhibits unwanted rotational slip. The object that is the source of the unsuccessful trial was the golf ball. During the execution of rolling the object rolled out of the fingers. This might be a result of the dimples reducing the total contact area made with the sensing surface. Fig.~\ref{fig:ball_recon} serves as visual verification of the 3D reconstruction.

\section{CONCLUSION}
\label{sec:conclusion}
Most applications of robotic manipulation rely on the use of suction or parallel grips due to the simplicity of their controls. This however comes at the cost of requiring high precision vision-based perception systems in order to complete a task. Further more, the use of such grippers limit applications to pick-and-place. Tasks that require the use of an object, such as cutting with a knife, require not only to pick up an object but to also reorient it. For aforementioned grippers this requires special rigs, two-arm manipulators, or interacting with the environment. These issues constrain robotic manipulation to structured environments. Dexterous manipulation is seen as a way forward, but most systems controlling dexterous hands rely heavily on vision, which compound the difficulty of the problem. While there are a variety dexterous manipulators equipped with tactile sensors, they are either too rigid or do not provide information. 

In this paper, we have presented a high-resolution, compliant, and round tactile sensor for dexterous manipulators. This required designing a new illumination system, that despite its limitations, is still suitable for manipulation. We justify this design by exploring the importance of the geometry of the sensor for dexterous manipulation. We also show that it is capable of being deployed into applications requiring real-time feedback by performing a set of in-hand manipulation in the form of controlled rolling on a diverse set of objects. 

Future work involves reducing the size of the sensor, making improvements in the illumination and 3D reconstruction, adding additional sensing modalities, and exploring the use of these sensors in more in-hand manipulation task and grasping.

\bibliographystyle{IEEEtran}
\bibliography{bibliography}

\end{document}